\documentclass[conference]{IEEEtran}
\IEEEoverridecommandlockouts

\newcommand{\pkg}[1]{\texttt{#1}}
\usepackage{cite}
\usepackage{hyperref}
\usepackage{amsmath,amssymb,amsfonts}
\usepackage{algorithmic}
\usepackage{graphicx}
\usepackage{svg}
\usepackage{textcomp}
\usepackage{xcolor}
\def\BibTeX{{\rm B\kern-.05em{\sc i\kern-.025em b}\kern-.08em
    T\kern-.1667em\lower.7ex\hbox{E}\kern-.125emX}}
\begin{document}

\title{An Empirical Evaluation of Time-Series Feature Sets}

\author{\IEEEauthorblockN{1\textsuperscript{st} Trent Henderson}
\IEEEauthorblockA{\textit{School of Physics} \\
\textit{The University of Sydney}\\
Sydney, Australia \\
then6675@uni.sydney.edu.au}
\and
\IEEEauthorblockN{2\textsuperscript{nd} Ben D. Fulcher}
\IEEEauthorblockA{\textit{School of Physics} \\
\textit{The University of Sydney}\\
Sydney, Australia \\
ben.fulcher@sydney.edu.au}
}

\maketitle

\begin{abstract}
Solving time-series problems using informative features has been rising in popularity due to the availability of numerous software packages for time-series feature extraction.
Feature-based time-series analysis can now be performed using any one of a range of time-series feature sets, including \pkg{hctsa} (7730 features: Matlab), \pkg{feasts} (42 features: R), \pkg{tsfeatures} (63 features: R), \pkg{Kats} (40 features: Python), \pkg{tsfresh} (up to 1558 features: Python), \pkg{TSFEL} (390 features: Python), and the C-coded \pkg{catch22} (22 features, able to be run from Matlab, R, Python, and Julia).
There is substantial overlap in the types of time-series analysis methods included in these feature sets (including properties of the autocorrelation function and Fourier power spectrum, and distributional shape statistics), but they are yet to be systematically compared.
Here we compare these seven feature sets on their computational speed, assess the redundancy of features contained in each set, and evaluate the overlap and redundancy across different feature sets.
We take an empirical approach to measuring feature similarity, based on the similarity of their outputs across a diverse set of real-world and model-simulated time series.
We find that feature sets vary across approximately three orders of magnitude in their computation time per feature on a laptop for a 1000-sample time series, from the fastest feature sets \pkg{catch22} and \pkg{TSFEL} ($\sim 0.1$\,ms per feature) to \pkg{tsfeatures} ($\sim 3$\,s per feature).
Using PCA to evaluate feature redundancy within each set, we find the highest within-set redundancy for \pkg{TSFEL} and \pkg{tsfresh}.
For example, in \pkg{TSFEL}, 90\% of the variance across 390 features can be captured with just four principal components.
Finally, we introduce a metric for quantifying overlap between pairs of feature sets, which indicates substantial overlap between the feature sets.
We found that the largest feature set, \pkg{hctsa}, is the most comprehensive, and that \pkg{tsfresh} is the most distinctive, due to its incorporation of large numbers of Fourier coefficients that are summarized at higher levels in the other sets.
Our results provide empirical understanding of the differences between existing feature sets, information that can be used to better understand and tailor feature sets to their applications.
\end{abstract}


\section{Introduction}

Time series, or ordered measurements taken over time, are a ubiquitous form of data in the sciences and industry.
Applications of time-series analysis are correspondingly diverse, from forecasting (the spread of disease through a population or future rainfall) to classification (from inferring cord pH from fetal heart-rate dynamics \cite{b1} to detecting abnormalities in silicon semiconductors \cite{b4}).
As a result, there is a wide range of quantitative tools for analyzing time series.

In this work, we focus on methods for extracting informative summary statistics, or \textit{features}, from univariate time series (sampled uniformly in time).
Such methods are diverse, and capture properties of the time-series distribution of values, autocorrelation structure, entropy, model-fit statistics, nonlinear time-series analysis, stationarity, and many others \cite{b6, b5}.
The first comprehensive effort to organize the interdisciplinary time-series analysis literature, using an approach termed \textit{highly comparative time-series analysis}, encoded thousands of diverse analysis methods as features and compared their behavior on a wide range of time-series data \cite{b5}.
This culminated in a software framework that includes implementations of over 7\,000 time-series features, as the Matlab package \pkg{hctsa} \cite{b6, b7}.
While \pkg{hctsa} is comprehensive in its coverage of time-series analysis methods, it involves a computational cost exceeding what is required of many real-world applications, for which a smaller number of simpler features is often sufficient.
Additionally, \pkg{hctsa} requires access to the proprietary Matlab software, which limits broader use.
To address this limitation, \pkg{catch22} is a reduced set of 22 canonical time-series features, that were derived from \pkg{hctsa} via a procedure designed to both maximize classification performance across 93 tasks and reduce redundancy among the top-performing features \cite{Lubba2019}.
These 22 features capture the main conceptual areas of time-series analysis covered in \pkg{hctsa}, including properties of distributional shape, linear and nonlinear autocorrelation, long-range scaling, symbolic transition rates, and others.
The \pkg{catch22} feature set is coded in C, with wrappers for Python and Matlab, and derived native packages in R (\pkg{Rcatch22} \cite{b10}) and Julia (\pkg{Catch22.jl} \cite{catch22jl}).

Other independently developed packages for computing time-series features also exist in R, including \pkg{feasts} \cite{b8} and \pkg{tsfeatures} \cite{Hyndman2020}.
\pkg{feasts} is an updated and refined iteration of \pkg{tsfeatures}, and both include sets of features tailored largely to time series commonly encountered in econometrics, such as seasonality and autoregressive processes.
Examples of features included in these sets include those associated with generalized autoregressive conditional heteroscedasticity (GARCH) models, crossing points, seasonality, and Seasonal and Trend decomposition using Loess (STL).

In Python, three popular libraries for time-series feature extraction are \pkg{tsfresh} \cite{b12}, \pkg{TSFEL} \cite{b13}, and \pkg{Kats} \cite{b14}. 
\pkg{tsfresh} is a popular feature set that contains implementations of up to 1\,558 features, including autocorrelation and Fourier decomposition methods, entropy, and distributional properties.
\pkg{TSFEL} includes features associated with autocorrelation properties and Fourier transforms, spectral quantities and wavelet decompositions, and distributional characteristics.
The feature set included in \pkg{Kats}, developed by Facebook's Engineering team, was designed to be similar in composition to R's existing \pkg{tsfeatures} package and includes features associated with crossing points, STL decomposition, sliding windows, autocorrelation properties, and Holt-Winters algorithms.

There are major barriers to leveraging these various feature-extraction packages, as they span programming languages and differ in required data formats and syntax.
For example, \pkg{Kats} requires a \texttt{DateTime} formatted time index and a corresponding numeric vector of values, whereas other packages represent uniformly sampled time-series data as an ordered vector of values.
The R package, \pkg{theft}: Tools for Handling Extraction of Features from Time series \cite{b15}, addresses these difficulties, providing a standardized computational framework for time-series feature extraction, supporting the \pkg{catch22}, \pkg{feasts}, \pkg{tsfeatures}, \pkg{tsfresh}, \pkg{TSFEL}, and \pkg{Kats} feature sets.

While the time-series analysis community now has ready access to multiple time-series feature sets, how does one select an appropriate feature set to use for a given application?
Does the behavior of features in different feature sets largely overlap, or are some distinctive?
Different applications may require different trade-offs between feature comprehensiveness and computation speed---how do the feature-sets compare?
While some sets contain many features, are these features largely capturing independent properties of the data, or are they highly correlated (and hence redundant), such that care will need to be taken to statistically account for the non-independence (or dimensionality reduction may be particularly useful).
Addressing these important questions requires systematic comparison of the behavior of the existing tools for time-series feature extraction, which has not previously been done.
In this work we aim to address this shortcoming by comparing seven time-series feature sets: \pkg{hctsa} (v1.06), \pkg{catch22} (v0.1.12), \pkg{feasts} (v0.2.1), \pkg{tsfeatures} (v1.0.2), \pkg{tsfresh} (v0.18.0), \pkg{TSFEL} (v0.1.4), and \pkg{Kats} (initial release).
We assess these feature sets in three different settings: (i) their computation time and its scaling with time-series length; (ii) the redundancy of features within a given feature set; and (iii) overlap and redundancy in feature behavior between feature sets.
We take a novel empirical approach of assessing features using their behavior across a diverse set of real-world and model-generated time series.
Note that \pkg{tsfeatures} includes multiple feature modules: we included all features here.

\section{Computation Time}

In applications involving large datasets, where computational resources are limited, smaller feature sets that are fast to compute are advantageous.
But for applications in which accuracy is prioritized, larger sets of complex time-series features may be slower to compute but can yield higher performance.
For example, some feature sets, like \pkg{hctsa}, contain complex nonlinear time-series analysis methods, and \pkg{tsfeatures} includes GARCH modeling.
These features are slower to compute than simple properties of the distribution (e.g., mean and variance) or linear correlation statistics (e.g., Fourier coefficients or linear autocorrelations).
It is thus important to empirically benchmark the computation-time scaling of different time-series feature sets with time-series length.

Here we estimated the computation-time scaling of each feature set across the length range of 100--1000 samples (indicative of time-series lengths in many applied problems).
We computed the time taken to evaluate each of the seven feature sets on a set of noise time series generated as independent samples from a Gaussian distribution, across five lengths, $T = 100, 250, 500, 750$, and $1000$.
The median and interquartile range of computation time was returned after repeating the process for ten independent noise time series of each length.
We confirmed that the results are not specific to our choice to use white noise time series for benchmarking; we found quantitatively similar results when repeating the analysis with noisy sinusoids (generated from: $x_t = 2 \sin(2t) + \epsilon$, for i.i.d. $\epsilon \sim \mathcal{N}(0,1)$ and $0 \leq t \leq 3\pi$).
Calculations were run on a 2019 MacBook Pro with an Intel Core i7 2.6 GHz 6-Core CPU, and all feature sets were timed in their native programming language (\pkg{hctsa} was run in parallel mode, allowing access to all six CPU cores).

Feature sets varied considerably in the number of features they contain, from \pkg{catch22} (22 features), \pkg{feasts} (43 features), \pkg{tsfeatures} (62 features), \pkg{Kats} (40 features), \pkg{TSFEL} (185 features for time series of length $T = 100$, 260 for length $T = 250$, 285 for length $T = 500$, and 390 for lengths $T > 500$; feature numbers scale approximately linearly with time-series length until $T > 500$), \pkg{tsfresh} (1\,558 features), and \pkg{hctsa} (7\,730 features).
All features computed successfully for all lengths for \pkg{catch22}, \pkg{feasts}, \pkg{tsfeatures}, and \pkg{TSFEL}.
For \pkg{tsfresh}, 392 features ($25.2$\%) did not compute successfully for $T = 100$, but no issues were noted for any other time-series lengths. 
These features all corresponded to fast Fourier transform (FFT) coefficients for real, imaginary, absolute, and angle components of 51--99. 
For \pkg{Kats}, 3 features ($7.5$\%) did not compute successfully across all time-series lengths.
These three features related to level, trend, and seasonal parameters of a Holt--Winters model fit, which understandably would not fit successfully on a white noise time series. 
For \pkg{hctsa}, 706 features ($0.91$\%) did not compute successfully for $T = 100$, 469 features ($0.61$\%) for $T = 250$, 458 features ($0.59$\%) for $T = 500$, 470 features ($0.61$\%) for $T = 750$, and 443 features ($0.57$\%) for $T = 1000$.
Some features (such as stationarity on a certain fixed timescale) will not compute on short time series, and some TISEAN-based features \cite{TISEAN}, requiring compilation from C and Fortran code, were excluded from this analysis.

\begin{figure*}
\centerline{\includegraphics[width=\linewidth]{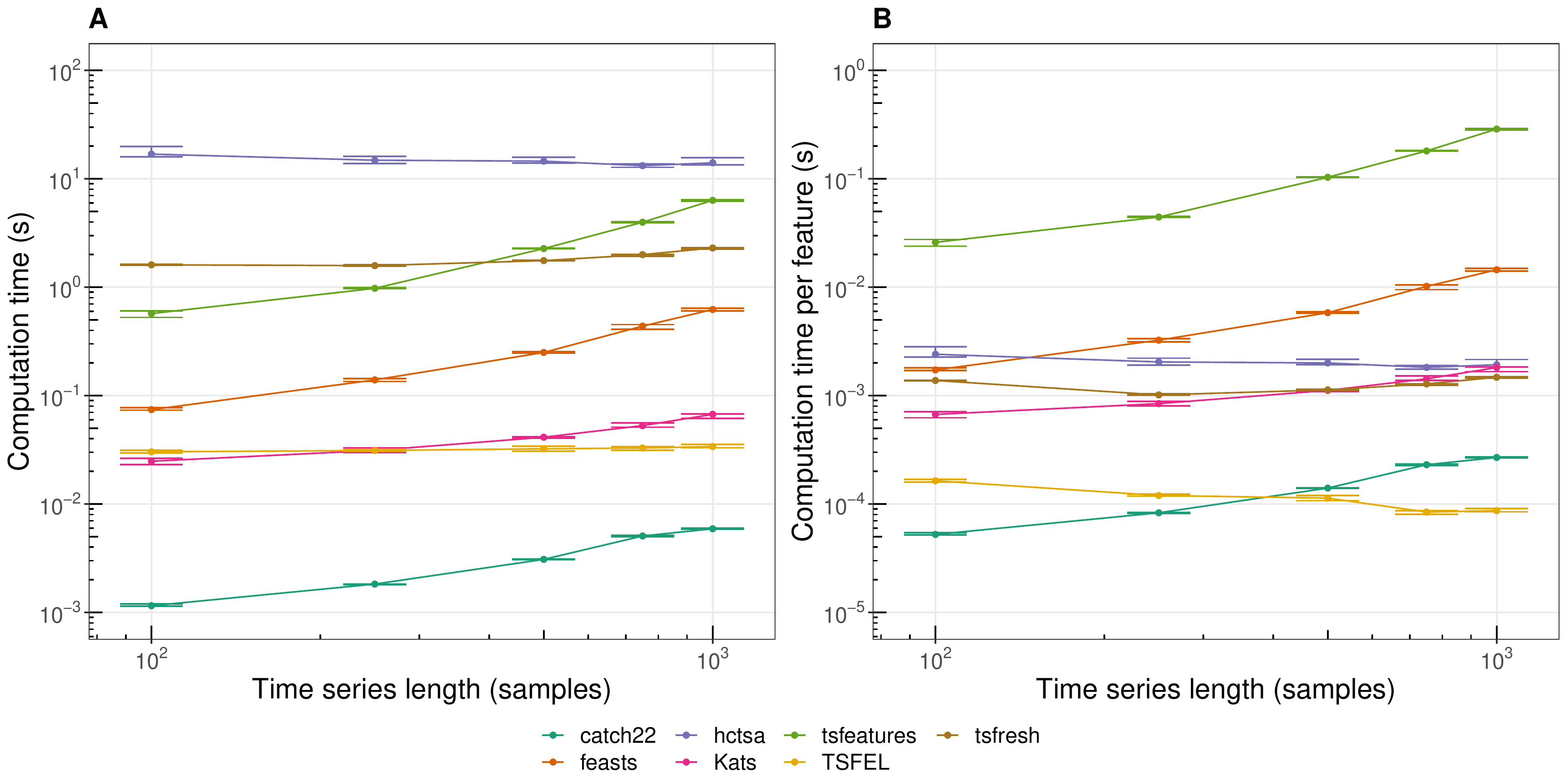}}
\caption{
\textbf{Computation times of different feature sets vary across orders of magnitude.}
\textbf{A} Median and interquartile range of computation time across ten Gaussian noise time series is plotted as a function of time-series length on logarithmic axes for lengths $T = 100, 250, 500, 750$, and $1000$.
\textbf{B} Median computation time per successfully computed feature is plotted for the same data as in \textbf{A}.
Results are shown for \pkg{catch22} (dark green), \pkg{feasts} (orange), \pkg{hctsa} (purple), \pkg{Kats} (pink), \pkg{tsfeatures} (light green), \pkg{TSFEL} (yellow), and \pkg{tsfresh} (brown).
}
\label{fig:comptime}
\end{figure*}

Median and interquartile ranges of computation times for each feature set are plotted as a function of time-series length in Fig.~\ref{fig:comptime}A.
We found that feature-set computation times varied over approximately four orders of magnitude.
For example, for a 1000-sample time series, feature-extraction times varied from $< 10$\,ms for the 22 C-coded features in \pkg{catch22}, to $16.5$\,s for the $7300$ \pkg{hctsa} features.
The other feature sets showed intermediate computation times, from \pkg{TSFEL} ($0.03$\,s) and \pkg{Kats} ($0.06$\,s), to \pkg{feasts} (0.47\,s), \pkg{tsfresh} (2.53\,s), and \pkg{tsfeatures} (6.18\,s).
To more fairly compare feature sets of different sizes (from the 22 features in \pkg{catch22} to the $>7000$ features in \pkg{hctsa}), we next investigated the computation time per successfully computed feature, shown in Fig.~\ref{fig:comptime}B.
We find that computation time per feature also varied over several orders of magnitude.
For a 1000-sample time series, feature computation times were fastest for \pkg{TSFEL} ($< 0.1$\,ms per feature) and \pkg{catch22} (0.3\,ms per feature); intermediate for \pkg{Kats} (2\,ms per feature), \pkg{hctsa} (2\,ms per feature), \pkg{tsfresh} (2\,ms per feature), and \pkg{feasts} (11\,ms per feature); and slowest for \pkg{tsfeatures} (280\,ms per feature).

\section{Evaluation Dataset}
\label{sec:evalDataset}

\begin{figure}
\centerline{\includegraphics[width=\linewidth]{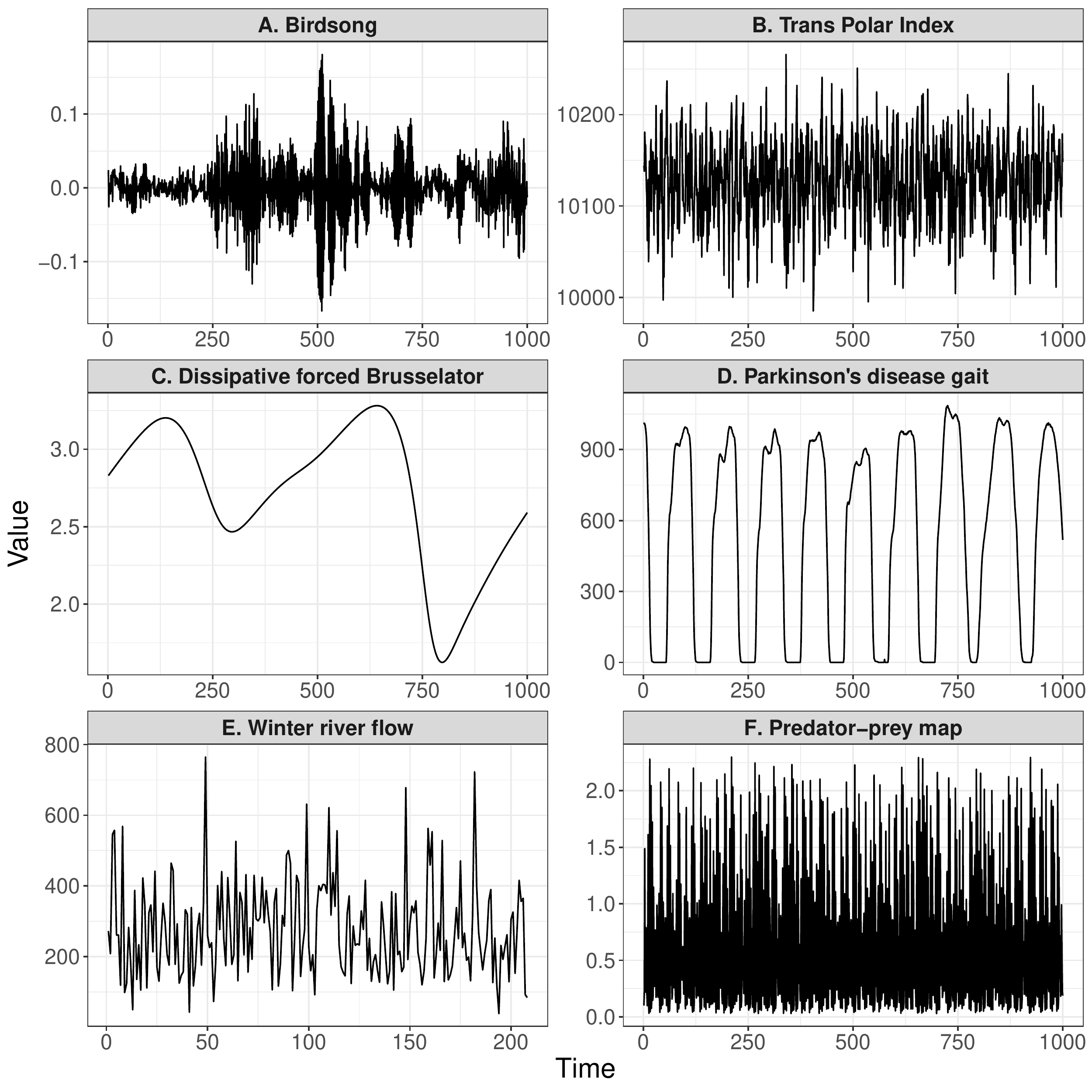}}
\caption{
\textbf{We assess time-series features based on their behavior on a dataset of 1000 diverse real-world and model-generated time series, a selection of which are plotted here \cite{Empirical1000, Fulcher2020CompEngine}.}
Six example time series from the Empirical 1000 dataset \cite{Empirical1000, Fulcher2020CompEngine} are plotted (after truncation to the first 1000 samples, as described in text):
\textbf{A} audio of the Black-bellied Plover (\texttt{AS\_s2.1\_f4\_b8\_l8280\_135971}),
\textbf{B} the Trans Polar Index normalized pressure difference between Hobart and Stanley
(\texttt{CM\_tpi\_hob}),
\textbf{C} a simulation from the dissipative forced Brusselator ODE \cite{b18} 
(\texttt{FL\_fbruss\_L300\_N10000\_IC\_0.3\_2\_y}),
\textbf{D} a gait time series from the Parkinson's disease gait dataset \cite{b19, b20}
(\texttt{MD\_gaitpdb\_GaCo05\_01l\_SNIP\_1256\-2955}),
\textbf{E} river flow
(\texttt{ME\_winter\_rf}), and
\textbf{F} a simulation from the dissipative predator-prey map \cite{b18}
(\texttt{MP\_predprey\_L5000\_IC\_0.5\_0.55\_x}).
Time-series names are provided in parentheses for reference to their corresponding data entries on \url{www.comp-engine.org} \cite{Fulcher2020CompEngine}.
}
\label{fig:emp1000}
\end{figure}

Having evaluated the differences in computation time of different feature sets, we next aimed to compare their behavior on real data, in order to assess the similarity of the constituent features, both within and between feature sets.
To do this, we use the `Empirical 1000' dataset (version 10) \cite{Empirical1000}, which consists of a diverse set of 1000 real-world and model-generated time series \cite{Fulcher2020CompEngine}.
Model-generated data spans a broad range of generative processes, including chaotic and non-chaotic deterministic dynamical systems (both discrete and continuous), and a range of stochastic processes.
The real-world time series span a broad range of domains, including hydrology, meteorology, finance, astrophysics, medicine, and animal-sound audio.
While the Empirical 1000 time series range in length from $T = 208$ to $T = 10\,000$ samples, we analyzed the first 1000 samples for time series longer than 1000 points to minimize the computational expense.
To demonstrate the range of temporal patterns contained in the Empirical 1000 dataset, six example time series (after truncation) are plotted in Fig.~\ref{fig:emp1000}.
This diversity of dynamical patterns reflects the corresponding diversity in the types of real-world and generative models studied in science and industry \cite{Fulcher2020CompEngine}.

Given the nature of some of the time series in the Empirical 1000, some individual features failed to compute successfully.
To minimize issues with further analysis, we employed a filtering process to retain the largest subset of features possible for each feature set while also removing consistently problematic time series from the Empirical 1000.
First, all features were first normalized independently using a $z$-score transformation.
We then filtered each set to only the $z$-score transformed features that had $<10$\% missing values.
This removed no features for \pkg{catch22} or \pkg{TSFEL}, 2 features ($4.7$\%) for \pkg{feasts} (41 remaining), 3 features ($7.5$\%) for \pkg{Kats} (37 remaining), 34 features ($4.4$\%) for \pkg{tsfresh} (745 remaining---a maximum of 779 features were computed on the Empirical 1000 compared to the 1\,558 features computed for the Gaussian noise computation time analysis), 3 features ($4.8$\%) for \pkg{tsfeatures} (59 remaining), and 22 features ($0.3$\%) for \pkg{hctsa} (7\,708 remaining).
After filtering for problematic features, we then filtered out any time series in the Empirical 1000 with missing values across any feature in any set.
This removed 105 time series, leaving 895 time series remaining.
Of the 105 removed problematic time series, 99 ($94.3$\%) were generated due to features in \pkg{TSFEL}, ($3.8$\%) were generated due to \pkg{tsfresh}, 1 ($0.6$\%) was generated due to \pkg{feasts}, and 1 ($0.6$\%) was generated due to \pkg{tsfeatures}.
We use these 895 time series and the retained features for each set to characterize feature behaviour throughout the remainder of this paper.

\section{Within Feature-Set Redundancy}

In constructing a feature set, choices need to be made in selecting the range of conceptually distinct types of time-series analysis algorithms to include, their parametric variations, and how to represent the method's outputs as one or more real-valued features.
For example, after computing the linear autocorrelation function, one could simply extract the raw values of the function at a chosen set of time-lags, $\tau$ (e.g., $\tau = 1, \dots, 40$), but could also capture different properties of the shape of the function (e.g., an exponential fit), or extract a timescale estimate, such as the function's first $1/e$ or zero crossing.
Similar choices need to be made when extracting features from the output of many other time-series analysis methods, including a Fourier or wavelet decomposition, the Lyapunov exponent spectrum, and many others.
In this process of coding the outputs of time-series analysis methods as real-valued features, some feature sets that take a comprehensive approach, in which many features are included and thus use multiple features to capture similar properties, are expected to have higher feature--feature redundancy.
Other feature sets may take a more restrained approach, aiming to include a smaller set of features that are less redundant with each other.
When doing statistical analysis with a feature set, it is important to understand the effective number of independent features it measures, especially when there is a high degree of collinearity between the included features.

We therefore aimed to investigate feature-set redundancy by assessing the range of distinct behaviors exhibited by features in each of the seven feature sets.
Within-set redundancy was measured by the ability of low-dimensional feature components to capture variance in the full feature set when applied to the Empirical 1000 dataset \cite{Empirical1000}.
A small number of low-dimensional components will explain a large amount of variance in a feature set that contains lots of redundant features (that produce highly correlated outputs across a diverse time-series dataset).
But feature sets containing linearly independent features that display distinctive behavior on time-series data will require many more components to explain a high percentage of variance across the full set of features.
Here we used the linear dimensionality-reduction method, principal components analysis (PCA), on the 895 time series $\times$ filtered feature matrix obtained from performing feature extraction for a given feature set on the 895 time series in the Empirical 1000 dataset.
Extracted principal components were ordered by their percentage of explained variance.

\begin{figure}[htbp]
\centerline{\includegraphics[width=\linewidth]{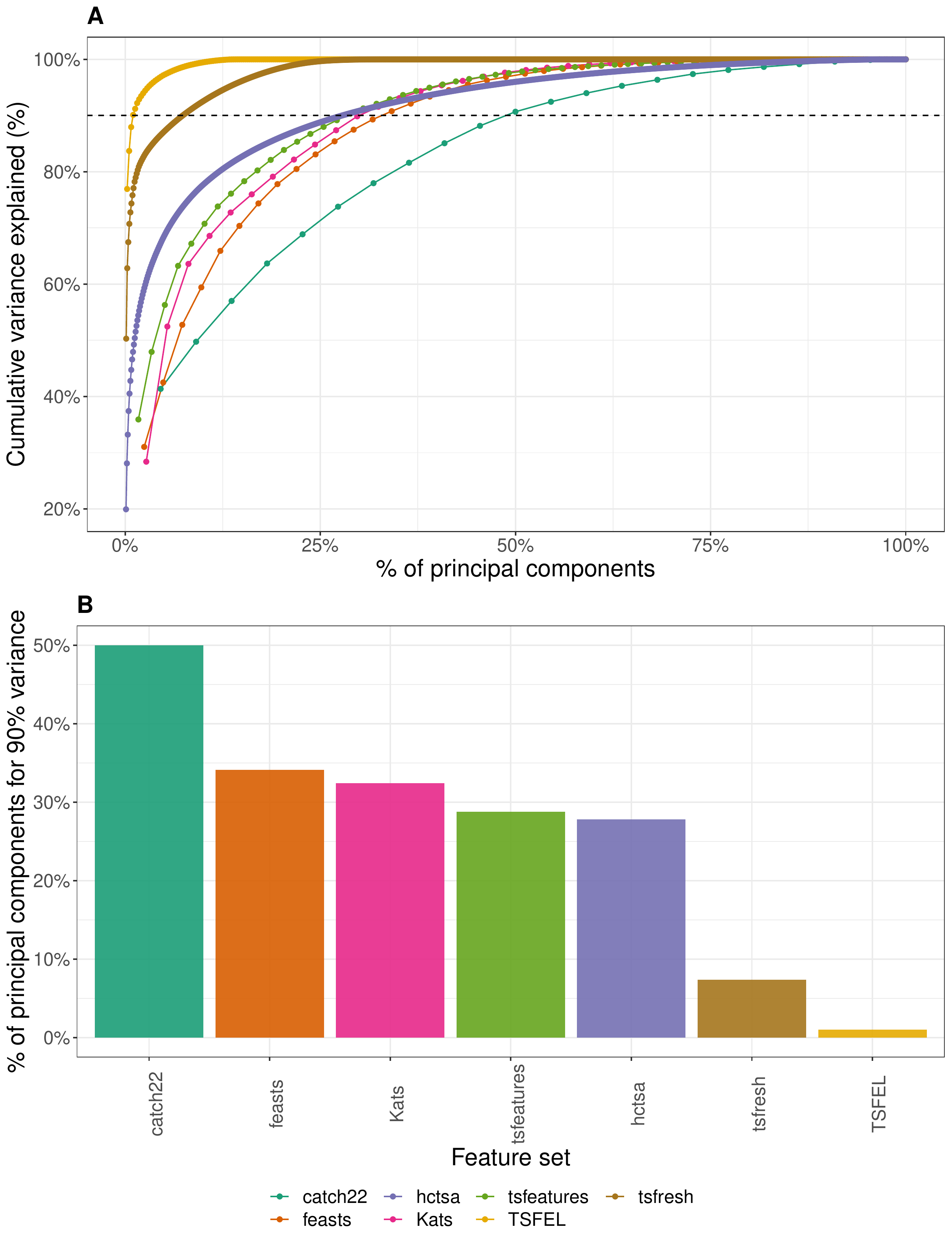}}
\caption{
\textbf{Feature sets vary substantially in their within-set redundancy.}
We assess within-set redundancy in terms of the ability of low-dimensional principal components to capture large amounts of variance in the feature space comprising a diverse set of 895 real-world and model-generated time series \cite{Empirical1000}.
\textbf{A} Cumulative variance explained is plotted as a function of the percentage of the number of principal components included for the seven labeled feature sets.
The dashed horizontal line indicates the 90\% cumulative variance threshold used in \textbf{B}.
\textbf{B} Percentage of the number of principal components required to explain 90\% of the variance in the feature space.
Results are plotted for the seven feature sets, ordered from least to most redundancy.
}
\label{fig:pca}
\end{figure}

To put feature sets of different sizes on a comparable scale, we expressed the number of principal components (PCs) as a proportion of the total number of PCs computed on the dataset.
This ranged from 22 for \pkg{catch22}, up to a maximum of 895 (the number of time series in the dataset).
We then summarized redundancy in a feature set using a simple summary statistic: the proportion of PCs required to explain 90\% of the variance in the feature space.
Results for this with full cumulative variance curves are plotted in Fig.~\ref{fig:pca}A and the overall summary statistic is plotted in Fig.~\ref{fig:pca}B.
The results reveal a substantial variation in feature redundancy across feature sets.
We observed the highest level of within-set redundancy for \pkg{TSFEL} and \pkg{tsfresh}.
For \pkg{TSFEL}, 90\% of the variance across all 390 features can be captured with just 4 PCs (1.0\%), and for \pkg{tsfresh}, 90\% of the variance of all 779 features was captured with 55 PCs (7.4\%).
To investigate this result, we inspected the features that make up these feature sets.
We found that a large proportion of features in both feature sets were derived from fast Fourier transform (FFT) coefficients: 256/390 (or 66\%) of \pkg{TSFEL} features, and 400/779 (or 51\%) of \pkg{tsfresh} features.
This large proportion of features derived from the FFT, in different forms of the raw coefficients, contributes to the high redundancy of features in \pkg{TSFEL} and \pkg{tsfresh}.

The most diverse feature set for its size was \pkg{catch22}, for which 11 PCs (50\%) were required to capture 90\% of the variance across its 22 features.
This is consistent with the construction of \pkg{catch22} as a reduction from 1000s of initial features that included a clustering procedure to reduce redundancy \cite{Lubba2019}.
An intermediate level of redundancy is seen in \pkg{feasts}, \pkg{tsfeatures}, and \pkg{Kats}, which have overlaps between their constituent features (the \pkg{Kats} feature set was based on \pkg{tsfeatures}, and \pkg{feasts} includes five of the same features as \pkg{tsfeatures}).
Despite its goal of being as extensive as possible in surveying time-series analysis algorithms, \pkg{hctsa} exhibited an intermediate level of redundancy, with 24.9\% of the PCs required to capture 90.0\% of the variance across all features.

\section{Between-Set Overlap and Redundancy}

Feature sets vary both in their coverage of time-series analysis methods and in their style of distilling these methods in the form of features.
Do some sets contain many unique features (that are not contained in other sets), or do some feature sets reproduce the behavior of features contained in other feature sets?
In this section, we aimed to address these questions by investigating the similarity between features contained in different feature sets.
We used the same set of well-behaved features and 895 time series that were used above (see Sec.~\ref{sec:evalDataset}): \pkg{catch22} (22 features), \pkg{TSFEL} (390), \pkg{feasts} (41), \pkg{Kats} (37), \pkg{tsfresh} (745), \pkg{tsfeatures} (59), and \pkg{hctsa} (7708).

We developed a method to quantify the redundancy of a feature set, given another feature set.
To achieve this for a given pair of feature sets, we first labeled one as a \textit{benchmark}, $B$.
We then aimed to capture how similar features in the second set, labeled \textit{test}, $T$, were to the features in the benchmark set.
If all individual features in the test feature set, $T$, behave similarly to a feature that is contained in the benchmark set, $B$, then we conclude that $T$ is redundant given $B$.
To quantify the similarity between two features, we used the absolute value of the Spearman correlation coefficient, $|\rho|$, between their outputs on 895 diverse time series from the Empirical 1000 dataset \cite{Empirical1000}.
In this way, features with highly correlated outputs on these diverse data are judged as having similar behavior.
We defined an ordered similarity metric, $\mathcal{S}$, to judge the redundancy of a feature set, $T$, given a benchmark set, $B$, based on the similarities of their constituent features as
\begin{equation}
    \label{eq:TestBenchmarkSim}
    \mathcal{S}(T|B) = \langle |\rho|^\mathrm{max}_i \rangle \,,
\end{equation}
where $|\rho|^\mathrm{max}_i = \max_j |\rho|_{ij}$ is the maximum Spearman rank correlation between each feature, $i$, in $T$, and all features, $j$, in $B$.
The average, $\langle \cdot \rangle$, is then taken over all features, $i$, in $T$, yielding $\mathcal{S}(T|B)$, the mean absolute maximum correlation between features in $T$, given those that exist in the benchmark, $B$.

\begin{figure}[htbp]
\centerline{\includegraphics[width=\linewidth]{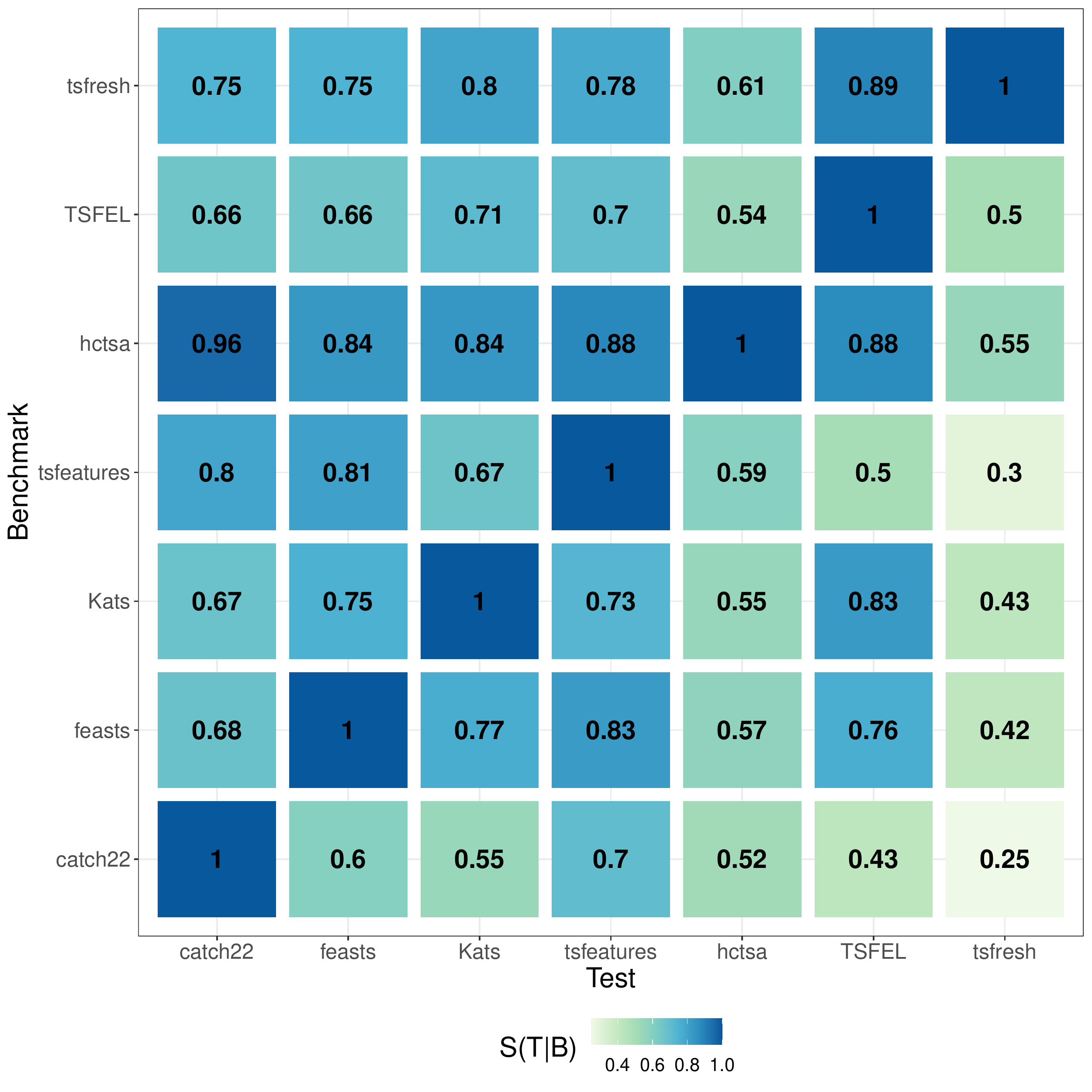}}
\caption{
\textbf{Time-series feature sets show a wide variation of overlap and redundancy.}
We assessed the redundancy of a given feature set by treating it as a `test' feature set, $T$, given another feature set, which is treated as a `benchmark', $B$.
The resulting directed similarity metric, $\mathcal{S}(T|B)$, Eq.~\eqref{eq:TestBenchmarkSim} (a mean of the maximum absolute Spearman correlation coefficients between feature outputs across a diverse time-series dataset), gives high values of $\mathcal{S}$ when features in $T$ reproduce the behavior of features that exist in $B$, and low values when features in $T$ are on average uncorrelated (distinctive) from the features in $B$.
Here we plot $\mathcal{S}(T|B)$ as a heatmap, where all feature sets have been treated as both a benchmark feature set (rows) and a test feature set (columns).
}
\label{fig:cor}
\end{figure}

Importantly, $\mathcal{S}$ is asymmetric, allowing us to understand the relative overlaps and redundancies between feature sets.
For example, consider a comprehensive feature set, $X$, containing many diverse and useful time-series features, and another simple set, $Y$, that only contains autocorrelation-related features.
Then $\mathcal{S}(Y|X)$ will be high (indicating that there is, on average, a feature in $X$ that is similar to each feature in $Y$), but $\mathcal{S}(X|Y)$ will be low (indicating that the simple features in $Y$ cannot capture the wider range of behavior of the features in $X$).
In the case that $X$ and $Y$ are sets containing very similar features, then both $\mathcal{S}(Y|X)$ and $\mathcal{S}(X|Y)$ will be high.
We computed $\mathcal{S}(T|B)$ for all pairs of feature sets, treating each feature set as both a benchmark and a test.
The result was represented as a pairwise matrix, with each benchmark feature set, $B$, shown as rows, and each test feature set, $T$, shown as columns, as shown in Fig.~\ref{fig:cor}.

The matrix reveals a range of interesting overlaps and redundancies between feature sets.
Overall, there are many high values, $\mathcal{S}$, indicating substantial overlap between the types of features included in the feature sets.
Looking across a row, feature sets with many high values, $\mathcal{S}$, are `comprehensive' in the sense that they contain features that exhibit similar behavior as those in the other feature sets.
As expected, \pkg{hctsa}, the package with the most numerous and comprehensive coverage of time-series analysis algorithms, was the strongest benchmark set: on average, there exists an feature in \pkg{hctsa} that is highly correlated to features in \pkg{catch22} ($\langle |\rho|^\mathrm{max}_i \rangle = 0.96$), \pkg{feasts} ($\langle |\rho|^\mathrm{max}_i \rangle = 0.84$), \pkg{Kats} ($\langle |\rho|^\mathrm{max}_i \rangle = 0.84$), \pkg{tsfeatures} ($\langle |\rho|^\mathrm{max}_i \rangle = 0.88$), and \pkg{TSFEL} ($\langle |\rho|^\mathrm{max}_i \rangle = 0.88$).
The exception is \pkg{tsfresh}, for which $\langle |\rho|^\mathrm{max}_i \rangle = 0.55$, indicating that, on average, for a feature selected from \pkg{hctsa}, the most correlated \pkg{tsfresh} feature will have $|\rho| = 0.55$.
Indeed, \pkg{tsfresh} is characterized by its distinctiveness overall: when considered as a test set, $T$ (inspecting the \pkg{tsfresh} column in Fig.~\ref{fig:cor}), $\mathcal{S}(\pkg{tsfresh}|B)$ values are even lower ($0.25-0.5$) for all other feature sets treated as benchmarks, $B$.
This indicates that \pkg{tsfresh} contains many unique time-series features that are not present in the other feature sets.

To investigate this uniqueness, we examined the individual features in \pkg{tsfresh} that were the least correlated to features in \pkg{hctsa}; in particular, those with $|\rho|^\mathrm{max} < 0.2$ (the maximum taken across all \pkg{hctsa} features).
We found 128 \pkg{tsfresh} features with $|\rho|^\mathrm{max} < 0.2$ to features in \pkg{hctsa}.
Of these, 118 (or 92\%) were Fourier coefficients (real and imaginary components, and angles).
While \pkg{hctsa} contains summary statistics derived from the Fourier power spectrum, they do not output the set of all raw coefficients from the fast Fourier transform (FFT).
This appears to drive the unique behavior of \pkg{tsfresh}, and also contributes to its relatively high within-set redundancy, cf. Fig.~\ref{fig:pca}.
Although \pkg{TSFEL} computes the absolute value of a large number of FFT coefficients, it does not also return the real and imaginary components and angles as features, and thus has a low $\mathcal{S}(\pkg{TSFEL}|\pkg{tsfresh}) = 0.5$.

As expected, given the history of their construction, \pkg{tsfeatures} and \pkg{feasts} were strongly overlapping in their behavior (in both directions, $\mathcal{S} \approx 0.82$), with high similarity scores also with the similarly constructed \pkg{Kats} ($\mathcal{S} \approx 0.72$).

\section{Discussion}

The feature-based approach to time-series analysis, in which a range of algorithms underpinned by theory can be used to distill a large time-series dataset down to a set of interpretable summary statistics is a promising way to address important time-series analysis problems in science and industry.
Here we characterized the behavior of, and overlaps between, the most prominent existing software libraries for time-series feature extraction: \pkg{hctsa}, \pkg{catch22}, \pkg{tsfeatures}, \pkg{feasts}, \pkg{Kats}, \pkg{tsfresh}, and \pkg{TSFEL}.
We took an empirical approach of characterizing features in terms of their pattern of outputs across a set of over 800 diverse real-world and model-generated time series, allowing us to distinguish `similar' features as those with highly correlated behavior on data.
We found that \pkg{TSFEL} and \pkg{catch22} were the fastest to evaluate for time series in the range 100--1000 samples ($\sim 10^{-4}$\,s per feature).
However, our low-dimensional analysis highlighted a large amount of redundancy in particularly \pkg{TSFEL} and \pkg{tsfresh}, with a large proportion of highly correlated features (including many raw Fourier coefficients) in both feature sets.
By contrast, \pkg{catch22}, a feature set designed to reduce within-set redundancy, displayed the least within-set redundancy.
We also analyzed the overlap between feature sets by introducing a directed pairwise metric of feature-set redundancy, that treats one set as a `benchmark' and measures how correlated features in a second `test' feature set are to those in the benchmark set.
As expected, \pkg{hctsa} was distinguished as the most comprehensive feature set, with features in \pkg{catch22}, \pkg{feasts}, \pkg{Kats}, \pkg{tsfeatures}, and \pkg{TSFEL} all having a highly correlated match ($\langle \rho \rangle \geq 0.84$) to a feature contained in \pkg{hctsa}.
The exception was \pkg{tsfresh}, which was marked as the most unique feature set, largely driven by a large number of features that capture the raw outputs of a fast Fourier transform as real and imaginary components, magnitudes, and phase angles (whereas many other packages extract summary statistics from the Fourier power spectrum).

Our results also highlight different choices made in feature composition and behavior across the seven time-series feature extraction libraries.
It is important to understand the characteristics of different feature sets, as different applications of time-series feature extraction call for different choices, such as prioritizing speed or accuracy.
Different feature sets take different approaches to feature-set construction, from the large \pkg{hctsa} feature set, which was developed with the aim of representing a comprehensive survey of the time-series analysis literature, to its distillation as the fast and less correlated features in \pkg{catch22}.
We also found a large amount of redundancy in many of the feature sets, such as \pkg{TSFEL} and \pkg{tsfresh}, where the variance across a large number of included features can be well captured using a small proportion ($<10$\%) of reduced components.
This redundancy was driven by large numbers of features quantifying similar types of time-series structure (such as Fourier coefficients or quantiles of the distribution of values).
Feature comprehensiveness comes at the cost of redundancy, and could be beneficial for some applications, e.g., involving large datasets and for which raw Fourier coefficients (including phase information of specific components) is relevant.
When applying statistical models to data, it is also important to be aware of and properly account for high collinearity of features within a feature set, which is substantial in all feature sets analyzed here.

While this work focused on the similarity of feature behavior on diverse time-series data (from the Empirical 1000 dataset \cite{Empirical1000, Fulcher2020CompEngine}), it will be important for future work to properly evaluate the relative performance of the different feature sets on different types of problems involving time series.
For example, the most common application of feature-based time-series analysis has been to time-series classification tasks---comparing the performance of different feature sets for different types of classification (and other) tasks will allow for clearer understanding of the relative strengths and weaknesses of these feature sets.
Other tasks which may yield informative differences between feature sets include large-scale anomaly detection \cite{anomalydetection} and feature-based forecasting \cite{forecasting}.
Our results may also depend on our choice to quantify feature similarity using a wide variety of data from the Empirical 1000 dataset \cite{Empirical1000}.
While features that are highly correlated across such diverse data can be considered to be measuring similar time-series properties in similar ways, feature--feature relationships may differ on more specific datasets, or on time series of lengths outside of the length range ($\leq 1000$) considered here.
We note that the R package, \pkg{theft} \cite{b15}, provides easy access to the six Python and R feature sets analyzed here (excluding the Matlab-based \pkg{hctsa}), allowing researchers to leverage combinations of features from different feature sets.
In the future, we hope that comparisons of time-series analysis methods across disciplines and feature sets will highlight fruitful directions for leveraging the combined strengths of diverse features, towards optimising an analysis approach for a given task.

\section{Available Code}

Code to reproduce all analysis included in this paper (including downloading and processing the Empirical 1000 dataset \cite{Empirical1000}) is available at \url{https://github.com/hendersontrent/feature-set-comp}.

\end{document}